\RequirePackage{fix-cm}
\documentclass[twocolumn]{svjour3}          
\smartqed  
\usepackage{cite}
\usepackage{epsfig}
\usepackage{graphicx}
\usepackage{mathptmx}      
\usepackage{latexsym}
\journalname{Nonlinear Dynamics}
\begin{document}

\title{Robustness of classification ability of spiking neural networks}


\author{ Jie Yang
\and Pingping Zhang
\and Yan Liu
}


\institute{Jie Yang \and Pingping Zhang\at
School of Mathematical Sciences, Dalian University of Technology, Dalian 116024, China.\\
\email{yangjiee@dlut.edu.cn; jssxzhpp@dlut.edu.cn}
\and
Yan Liu\at
School of information Science and Engineering, Dalian Polytechnic University, Dalian 116034, China.\\
\email{liuyan.3001@gmail.com}           
}

\date{Received: 15 Jul 2014 / Accepted: 29 May 2015}

\maketitle

\begin{abstract}
It is well-known that the robustness of artificial neural networks (ANNs) is important for their wide ranges of applications.
In this paper, we focus on the robustness of the classification ability of a spiking neural network which receives perturbed inputs.
Actually, the perturbation is allowed to be arbitrary styles.
However, Gaussian perturbation and other regular ones have been rarely investigated. 
For classification problems, the closer to the desired point, the more perturbed points there are in the input space.
In addition, the perturbation may be periodic. 
Based on these facts, we only consider sinusoidal and Gaussian perturbations in this paper. 
With the SpikeProp algorithm, we perform extensive experiments on the classical XOR problem and other three benchmark datasets.
The numerical results show that there is not significant reduction in the classification ability of the network if the input signals are subject to sinusoidal and Gaussian perturbations.
\keywords{Robustness \and Spiking neural networks \and Gaussian perturbation \and Classification}
\end{abstract}
\section{Introduction}
Lots of biological experiments and theoretical analysis have demonstrated that the speed and scale of processing information by biological neural networks are much faster and larger than by manual methods~\cite{1}\cite{2}. 
Inspired by animals' central nervous systems in particular the brain, many kinds of artificial neural networks (ANNs) and training methods have presented to mimic animals' behavior characteristics. 
ANNs are distributed mathematical models that process information parallelly\cite{3}\cite{4}.
They have been used to solve a wide variety of tasks that are hard to solve by using ordinary rule-based programming, including computer vision and speech recognition~\cite{5}\cite{6}\cite{7}\cite{8}. 
These networks base on the characteristics and scales of data and the complexity of systems. 
By adaptively adjusting the weights which are connected between different nodes in adjacent layers, ANNs can achieve the purpose of processing information.
As a special class of ANNs, spiking neural networks (SNNs) can simulate spikes generated between animal dendrites and axons of neurons \cite{4}\cite{9}. 
With temporal information coding in single spikes to process information, they were proved to be a type of strong anthropomorphic networks.

However, due to many uncontrollable factors, such as noising inputs, individual spike decay times, thresholding weights, the abilities of processing information by spiking neural networks may be affected~\cite{10}\cite{11}.
In order to get an available network architecture, it is important to do some research on its robustness. 
Because of encoding input variables by time differences between pulses, spiking neurons are particularly sensitive to the input signals.
Recently, this has led to several explorations on the computational abilities and learning performance of neuromorphic networks of spiking neurons with noise\cite{12}\cite{13}. 
However, these works have not considered the type of perturbations and the robustness of classification ability of SNNs.

To analyze the robustness of SNNs' classification abilities, in this paper we performed a series of numerical experiments on the classical XOR problem and other three benchmark datasets (\emph{i.e.}, Iris dataset, Wisconsin breast cancer dataset and StatLog landsat dataset) with the SpikeProp algorithm\cite{14}.
Notably, the closer the perturbed inputs to the desired points, the more perturbed points there are in the input space.
What's more, the perturbation may be periodic in practice. 
These facts led us to consider the sinusoidal and Gaussian perturbations in this paper.

To summarize, our main contributions include:
\begin{itemize}
\item
As far as we know, this is the first work to validate the robustness of classification ability of SNNs which receive perturbed inputs.
\item
Two kinds of perturbations were considered for robustness of classification ability of a SNN. In fact, perturbations could be arbitrary styles performing on the inputs. We only focus on sinusoidal and Gaussian perturbations.
\item
On the classical XOR problem and other three benchmark datasets, we evaluate the classification ability of SNNs and show its robustness experimentally.
\end{itemize}
\section{Spiking neural networks}
Compared with traditional neural networks (such as BP), SNNs have several differences in network architectures.
The most important one is that there are multiple synaptic terminals and the specific synaptic delay between spiking neurons in adjacent layers.
In addition, due to the fast temporal encoding which is very different from traditional rate-coded networks, spiking neurons can significantly improve complex non-linear classification performances~\cite{4}\cite{14}.
\subsection{The architectures of SNNs}
A simple feed-forward SNN with multiple input spiking neurons and one output spiking neuron is shown in Fig 1. 
The network architecture consists of one input layer, one hidden layer and one output layer, denoted by I, H and O respectively.
\begin{figure}
\begin{center}
\includegraphics[width=3in]{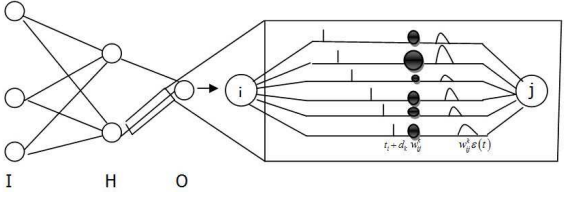}
\caption{The structure of a simple spiking neural network.}
\label{fig:1}
\end{center}
\end{figure}
Each connection between different layers comprises several synapses and each neuron receives a set of spikes from all its previous neurons. Formally, assuming that during the simulation interval each neuron generates at most one spike and fires when the internal state variable reaches a threshold, the state variable ${x_{j}}$ of neuron $j$ receives outputs from all its previous neurons as a weighted sum of the pre-synaptic contributions:
\begin{equation}
{x_{j}(t)=\Sigma_{i\in D_{j}}\Sigma_{k=1}^{m}w_{ij}^{k}y_{i}^{k}(t)}
\end{equation}
where $D_{j}$ denotes the set of pre-synaptic neurons associated with neuron $j$, $w_{ij}$ is the weight associated with synaptic terminal $k$, and $y_{i}^{k}(t)$ represents a delayed pre-synaptic potential (PSP) for each terminal,
\begin{equation}
{y_{i}^{k}(t)=\epsilon(t-t_{i}-d^{k})}
\end{equation}
with $\epsilon(t)$ a spike-response function. The time $t_{i}$ is the firing time of pre-synaptic neuron $i$, and $d^{k}$ is the delay associated with the synaptic terminal $k$. The firing time is determined as the first time when the state variable reaches the threshold. The spike-response function is always described as the form
\begin{equation}
\epsilon(t)= \left\{ \begin{array}{ll}
\frac{t}{\tau}\exp^{(1-\frac{t}{\tau})},& \textrm{$t>0$}\\
0 ,& \textrm{$t\leq0$}
\end{array} \right.
\end{equation}
where $\tau$ models the membrane potential decay time constant that determines the rise and decay time of the PSP.
\subsection{Learning algorithm}
\label{sec:2.2}
The basic SpikeProp algorithm\cite{14} is performed and we choose the least mean square as the error-function. Given desired spike times $\{t_{j}^{d}\}$ and actual firing times $\{t_{j}\}$, we can derive the form of the error-function
\begin{equation}
{E=\frac{1}{2}\Sigma_{j}(t_{j}-t_{j}^{d})^2}.
\end{equation}
For error back-propagation, the weights update rule is followed:
\begin{equation}
\mathnormal{w_{ij}^{k+1}(t_{j})=w_{ij}^{k}(t_{j})+\Delta w_{ij}^{k}(t_{j})}
\end{equation}
Define
 \begin{equation}
\delta_{j}=\frac{t_{j}^{d}-t_{j}}{\Sigma_{il}w_{ij}^{l}\frac{\partial y_{i}^{l}(t_{j})}{\partial t_{j}}}
\end{equation}
In the output layer, the basic weight adaptation function for neurons is derived as
\begin{equation}
\Delta w_{ij}^{k}( t_{j})
=-\eta\frac{y_{i}^{k}(t_{j})(t_{j}^{d}-t_{j})}{\Sigma_{il}w_{ij}^{l}\frac{\partial y_{i}^{l}(t_{j})}{\partial t_{j}}}=-\eta y_{i}^{k}(t_{j})\delta_{j}.
\end{equation}
For the hidden layers, the weight adaptation function for neurons is given by
\begin{equation}
\Delta w_{hi}^{k}(t_{j})
=-\eta\frac{y_{h}^{k}(t_{j})\Sigma_{j}\{\delta_{j}\Sigma_{k}w_{ij}^{k}\frac{\partial y_{i}^{k}(t_{j})}{\partial t_{j}}\}}{\Sigma_{nl}w_{ni}^{l}\frac{\partial y_{n}^{l}(t_{j})}{\partial t_{j}}}
\end{equation}
where $\eta$ is the learning rate of the network. For more details, see\cite{14}\cite{15}\cite{16}.
\section{Perturbations to neural networks}
\subsection{Traditional neural network perturbation approaches}
\label{sec:3.1}
Traditional neural networks are led to suitable Lyapunov-Krasovskii functional and by introducing appropriate random variables, such as the free weights techniques, one can analyze stochastic neural networks associated with the value of parameter uncertainties and numerical experiments to demonstrate the robust global exponential stability or asymptotic stability. In\cite{17}, the effect of both variation range and distribution of the time delay were taken into account for delay-distribution-dependent state estimation. The stochastic perturbations are described in terms of a Brownian motion and the time-varying delay is characterized by introducing a Bernoulli stochastic variable. In\cite{18}, some parameter matrices were used to express the relationships among the system variables which were perturbed to study the asymptotic stability of delay nonlinear cellular neural network. By perturbing the time variable interval of Hopfield neural networks, \cite{19} investigates the existence of the equilibrium points and global robust exponential stability .
\subsection{Our perturbation approach}
\label{sec:3.2}
 SNNs use the reacted pulses to transmit information. When the input signal enter into a spiking neural network, the state of each neuron will change (see formulas (1),(2),(3)). Once the state variable exceeds the threshold value, the neuron arouses a pulse. Hence input signals of SNNs have a great relevance to their robustness. A large number of practical examples imply that the signals after random perturbations will not be very different from the original one, and only few perturbed signals have large deviations from the original data. With these basic facts, we are motivated to perform different types of perturbations on input signals, and investigate the the robust classification abilities of SNNs.
 \par For the sake of simplicity, we only describe perturbations on XOR problem. For other three benchmarks, the operation is similar. Classical XOR problem requires hidden units to transform the inputs into the desired outputs and needs to classify 4 points (described as (0,0),(1,1),(1,0),(0,1) ) into 2 categories. It should be noted: if the four points are perturbed with a same value, it is equivalent to translating them simultaneously, leading no difference between the original inputs in essence. For convenience, we choose the point (1,1) as a target point to test the robustness of spiking neural networks. The basic perturbation formula is
\begin{equation}
{\tilde{x}=\tilde{x_{0}}+\sigma}
\end{equation}
with $\tilde{x_{0}}$ being the original input, and $\sigma$ the noise term. The formula means if $\tilde{x_{0}}=(a,b)$, the perturbed input $\tilde{x}$ will be $(a+\sigma,b+\sigma)$.
\par(1)Sinusoidal perturbations
\\The original input $\tilde{x_{0}}$ is perturbed by a sinusoidal perturbation term, which can be expressed as
\begin{equation}
{\tilde{x}=\tilde{x_{0}}+A\sin(2\pi y)}
\end{equation}
where $A$ is a constant between (0,1] to control the perturbation amplitude, and $y$ is a random vector and its component values belong to [0,1]. The numbers of components associated with $\tilde{x_{0}}$ and $y$ are the same.
\par(2)Gaussian perturbations
\\The original input $\tilde{x_{0}}$ is perturbed by a Gaussian perturbation. This is different from the sinusoidal perturbation, and can be expressed as
\begin{equation}
{\tilde{x}=\tilde{x_{0}}+\tilde{x_{0}}(I-\exp(-r^2/2)\cdot sgn(l))}
\end{equation}
where $I$ is an identity matrix, $r$ is a random vector whose component values belong to [0,1] and $sgn(l)$ is the signal function associated with $l$. Here the numbers of components associated with $\tilde{x_{0}}$, $r$ and $l$ are also the same.
\section{Experimental results}
Since our aim is to assess the robustness of classification ability of SNNs, we don't need to design complex SNNs. We train a simple spiking neural network on XOR problem and other three benchmark datasets.
\subsection{XOR problem}
\label{sec:4.1}
Firstly, the input and output signals of spiking neural networks are coded as in \cite{14}. If $Max$ and $Min$ are extremal values of a variable $x$ (e.g.an input signal), we can encode it as a spike fired in the time
\begin{equation}
\mathnormal{f(x)=\frac{x-Min}{Max-Min}\cdot L}
\end{equation}
with the length of coding interval $L$.
\par For a better representation of spike-time patterns, we can associate a 0 with a ``late'' firing time and a 1 with an ``early" firing time. With specific values 0 and 6 for the respective input times, we lead to the temporally encoded XOR\cite{14}\cite{20} showing in Table 1.
\begin{table}
\begin{center}
\caption{Encoded inputs and outputs for XOR problem}
\label{tab:1}       
\begin{tabular}{lll}
\hline\noalign{\smallskip}
Input &patterns  & Output patterns  \\
\noalign{\smallskip}\hline\noalign{\smallskip}
0 & 0 & 16 \\
0 & 6 & 10 \\
6 & 0 & 10 \\
6 & 6 & 16 \\
\noalign{\smallskip}\hline
\end{tabular}
\end{center}
\end{table}
In the table, the input numbers represent spike times (i.e. late firing times and early firing times) in milliseconds. The actual input patterns contain a setting threshold by adding a third input neuron. Here we define the difference between the times equivalent with ``0" and ``1" as the coding interval $L=6$ ms. We use the feed-forward network with connections that have a delay interval of 15 ms, so the available synaptic delays are 1-16 ms for classification. According to the formula (3), we calculated the postsynaptic states and selected $\tau=7$ ms.
\par Based on the above settings, numerical experiments were performed with a spiking neural network composed of one input layer, one hidden layer and one output layer. There were three input layer neurons (two encoding neurons and one threshold neuron), five hidden layer neurons (of which one inhibitory neuron generating only negative signal PSPs) and one output layer neuron. A weight and corresponding terminals ($m=16$) were configurated between each pair of synaptic neurons in adjacent layers.
\par Using computer simulations, we randomly generated 400 perturbed samples for different $r$ to see the distribution of perturbed samples as shown in Fig 2. Considering the computational complexity and time, we only used 160 perturbed samples for testing. We varied $A$ and $r^{*}$ (the least upper bound of all components to the random vector $r$ )to control perturbation amplitudes. The network reliably learned the XOR patterns with $\eta=0.01$. With different perturbations, we got the correct classification rate of the spiking neural network. The average of correct classification rates with and without perturbed data are compared in Table 2 and Table 3.
\begin{figure*}
\begin{center}
\scalebox{0.75}{\includegraphics[width=3in]{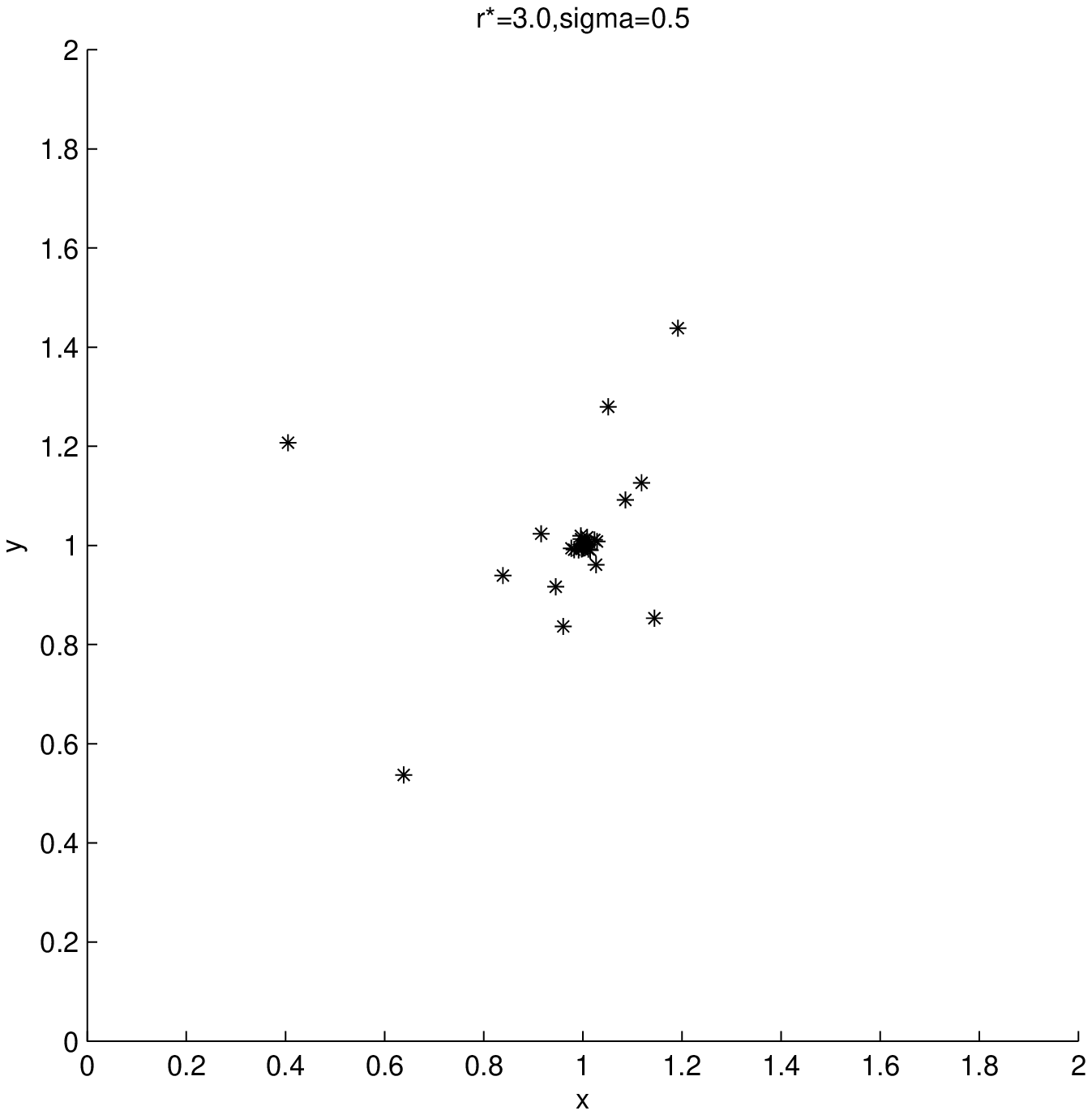}}
\scalebox{0.75}{\includegraphics[width=3in]{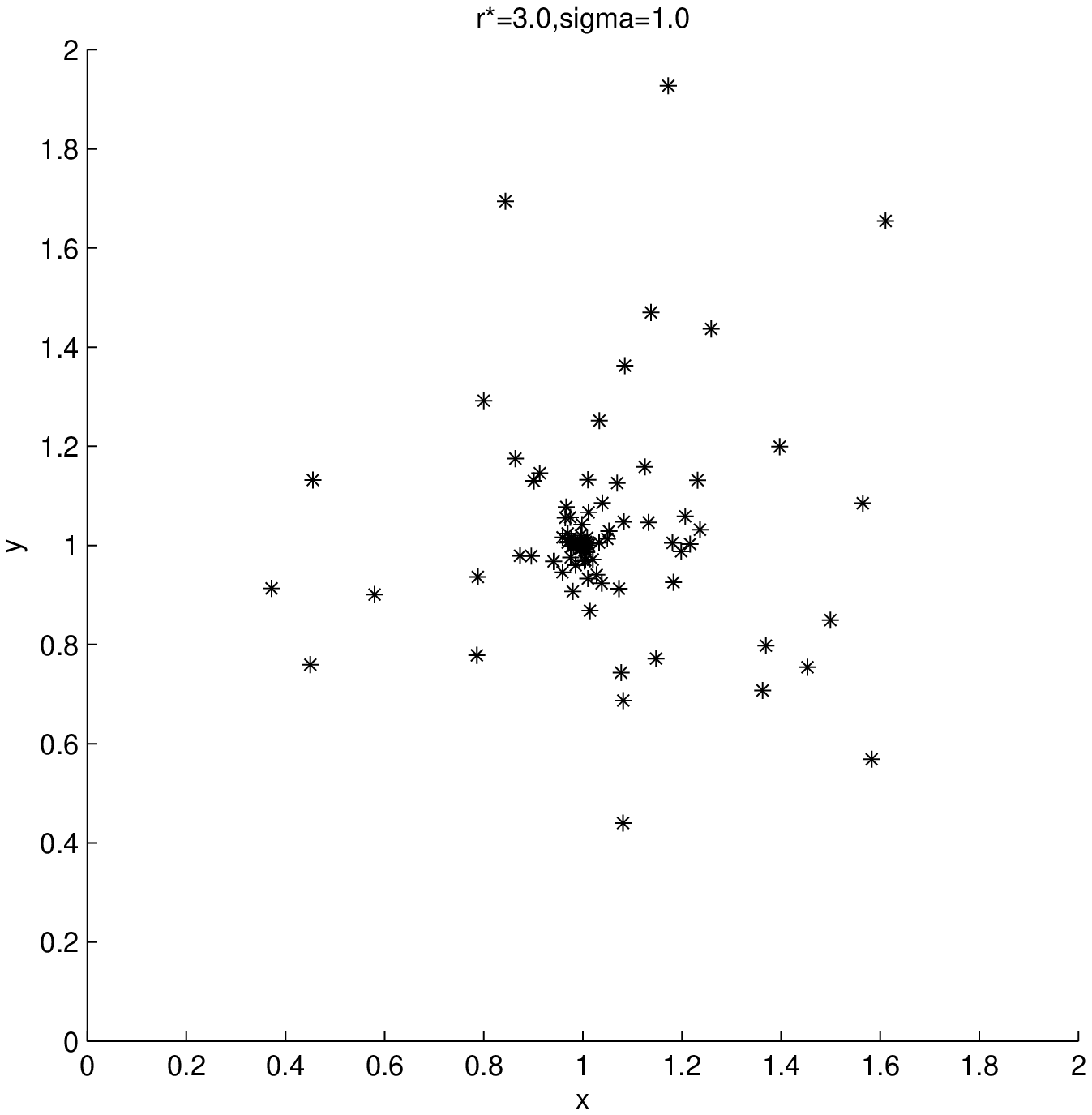}}
\scalebox{0.75}{\includegraphics[width=3in]{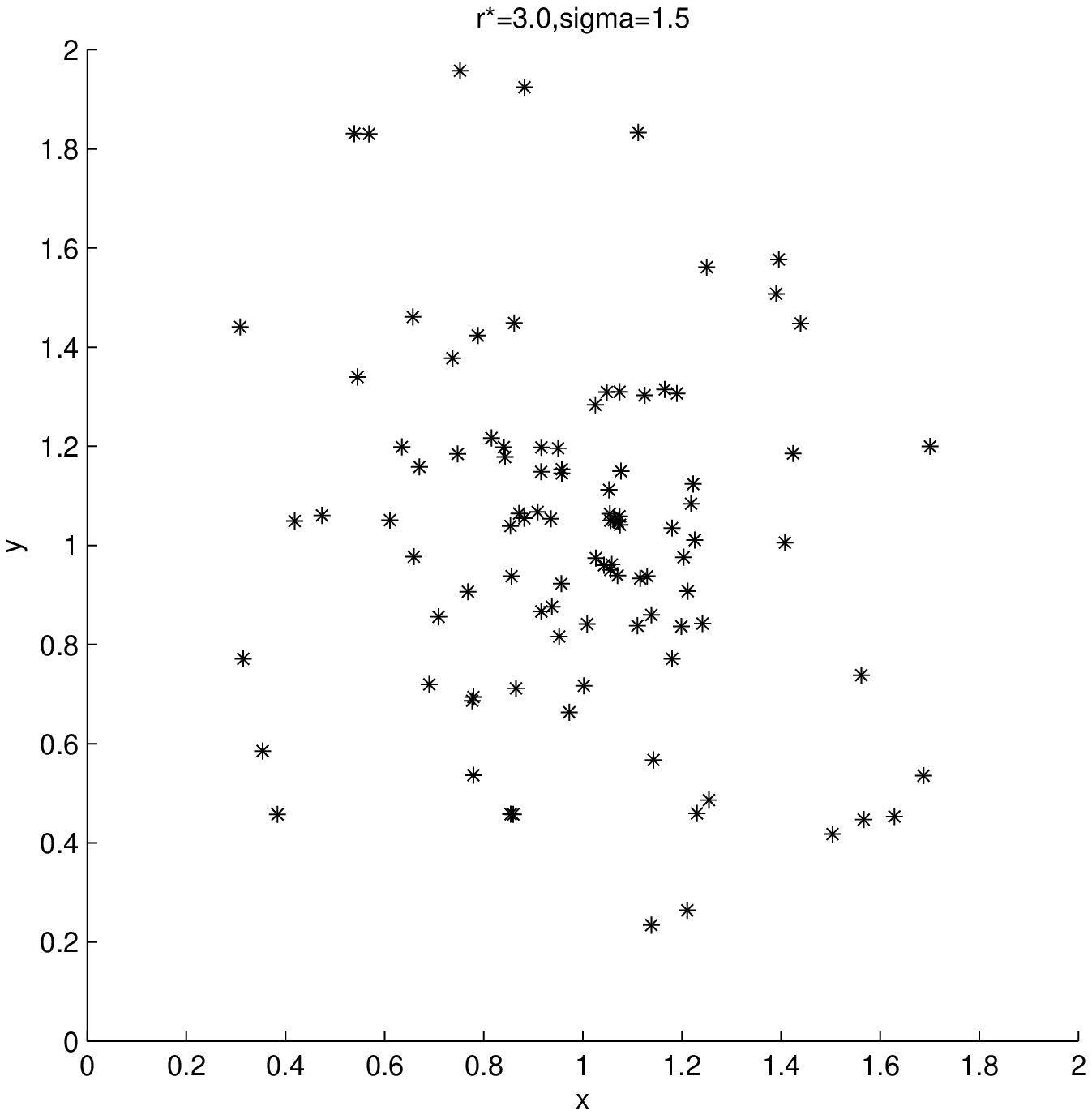}}
\scalebox{0.75}{\includegraphics[width=3in]{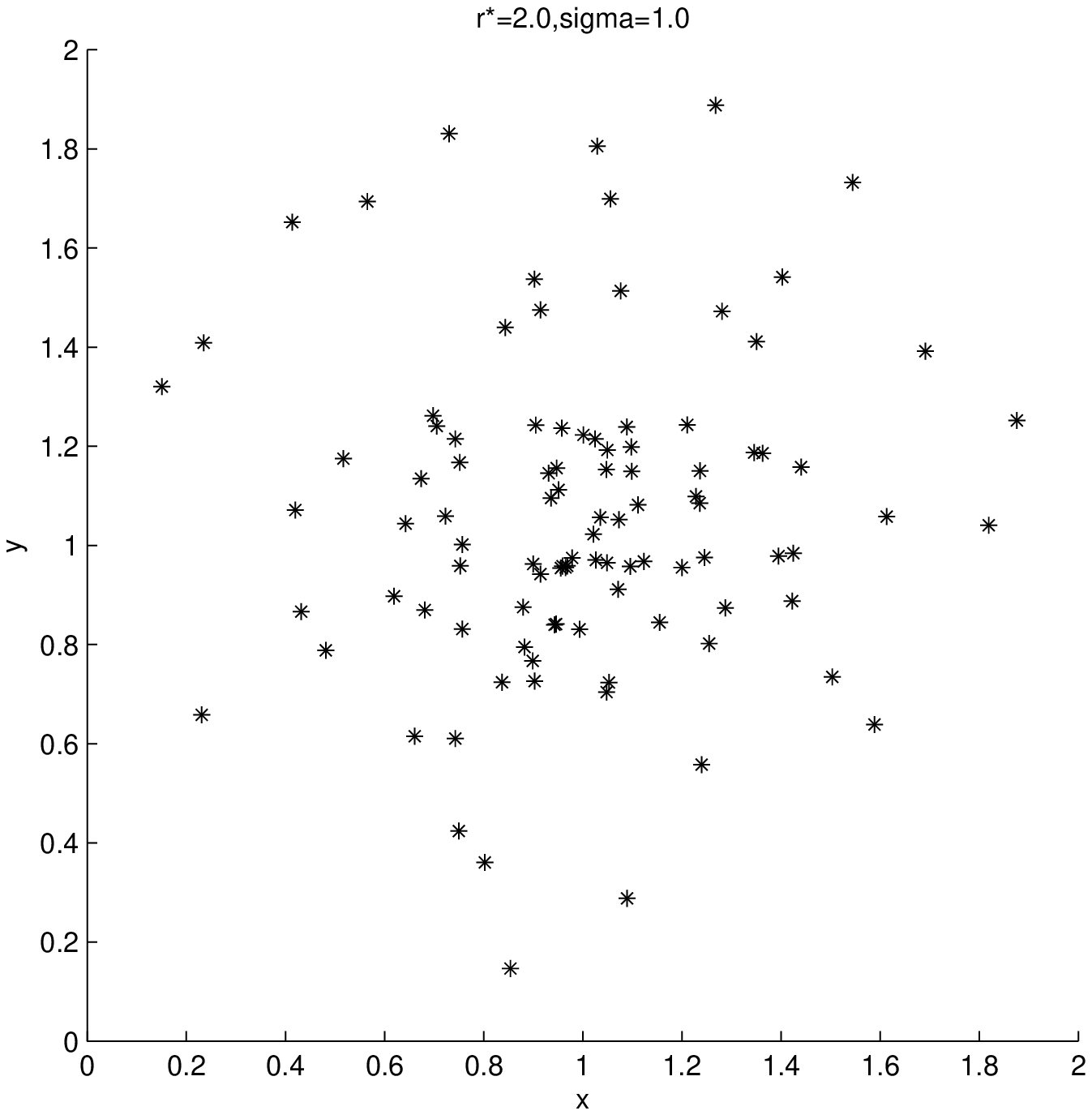}}
\scalebox{0.75}{\includegraphics[width=3in]{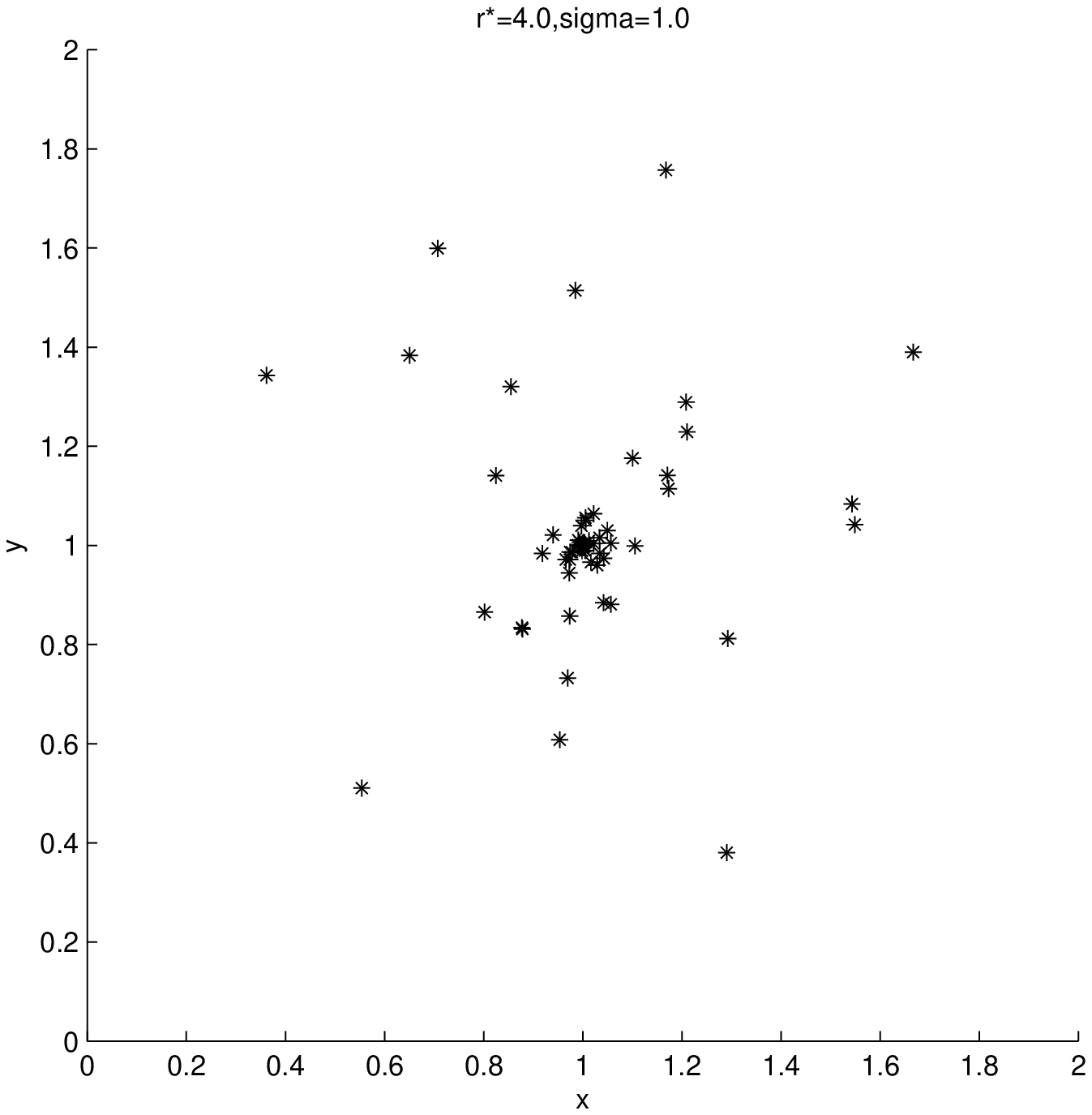}}
\scalebox{0.75}{\includegraphics[width=3in]{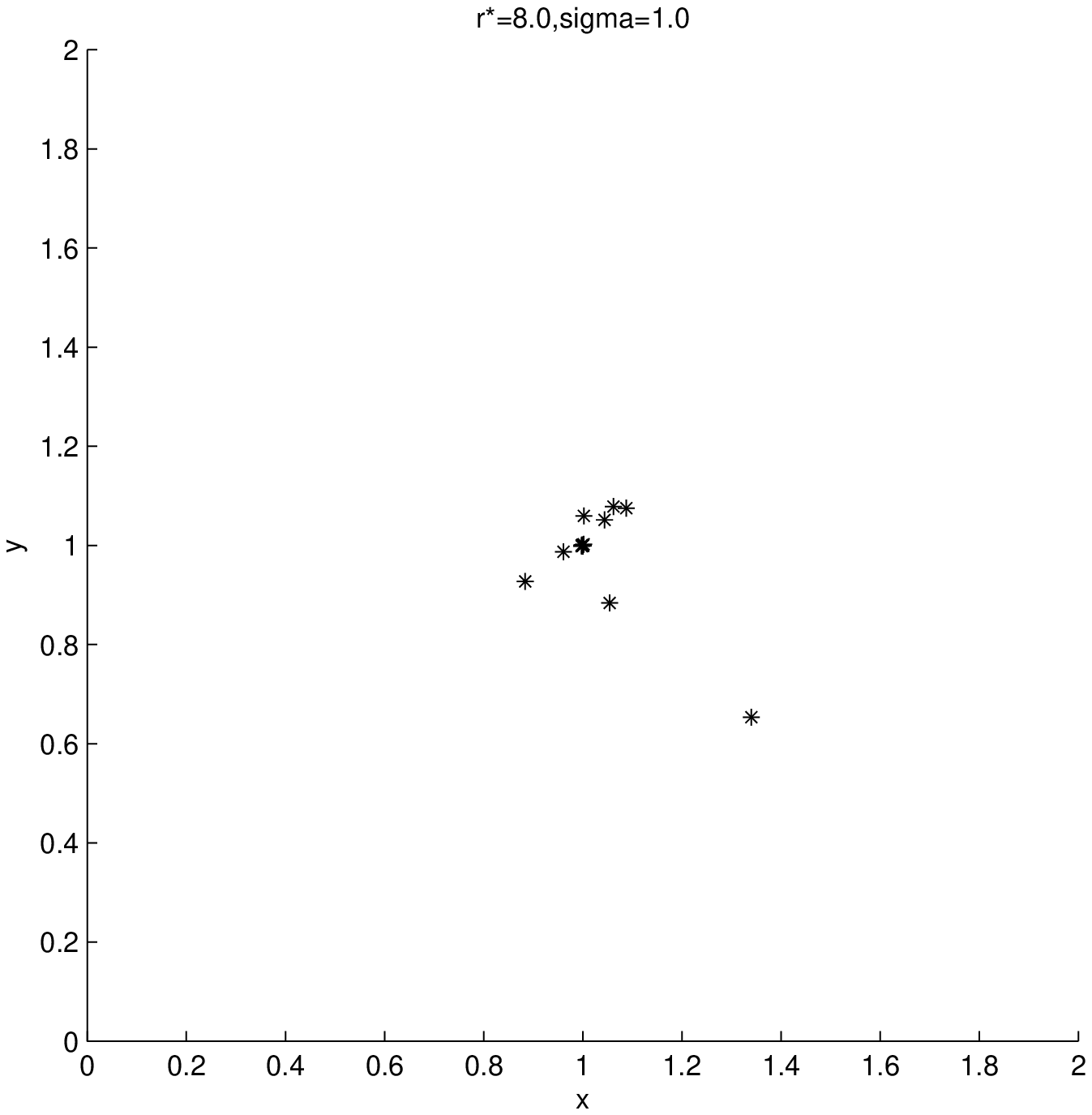}}
\caption{The perturbed samples for different $r*$. The desired point is (1,1) and the closer to the desired point, the more perturbed points there are. Besides, the perturbed points become more scattered when $r*$ increases.}
\label{fig:2}       
\end{center}
\end{figure*}

\begin{table}
\begin{center}
\caption{The result of the correct classification with sinusoidal perturbations. ROS: Rates of the correct classification without sinusoidal perturbations, RWS: Rates of the correct classification with sinusoidal perturbations.}
\label{tab:2}       
\begin{tabular}{llll}
\hline\noalign{\smallskip}
Epoches & A & ROS & RWS  \\
\noalign{\smallskip}\hline\noalign{\smallskip}
50& 0.001 & 90.50 & 91.00 \\
50& 0.01 & 89.50 & 87.80 \\
50& 0.1 & 91.00 & 88.90\\
50& 0.2 & 88.50 & 87.20 \\
50& 0.5 & 87.50 & 82.24\\
50& 0.8 & 87.50& 85.58 \\
\noalign{\smallskip}\hline
\end{tabular}
\end{center}
\end{table}
\begin{table}
\begin{center}
\caption{The result of the correct classification with Gaussian perturbations. ROG: Rates of the correct classification without Gaussian perturbations, RWG: Rates of the correct classification with Gaussian perturbations.}
\label{tab:3}       
\begin{tabular}{lllll}
\hline\noalign{\smallskip}
Epoches & $r^{*}$ & $\sigma$ & ROG & RWG  \\
\noalign{\smallskip}\hline\noalign{\smallskip}
50& 0.1 & 92.00 & 88.62 \\
50& 0.2 & 92.00 & 88.45 \\
50& 0.3 & 89.50 & 87.80\\
50& 0.4 & 89.50 & 88.15\\
50& 0.5 & 88.50 & 85.00\\
100& 0.1 & 88.25 & 88.66 \\
100& 0.2 & 89.75 & 89.60 \\
100& 0.3 & 91.75 & 90.86\\
100& 0.4 & 91.00 & 89.76\\
100& 0.5 & 88.25 & 88.81 \\
\noalign{\smallskip}\hline
\end{tabular}
\end{center}
\end{table}
\par As results are reported in Table 2 and Table 3, the classification accuracies associated with original data are different but almost equal (about 90\%).This is mainly because the input data were reordered after each epoch. The larger perturbation was performed to the input data, the more greatly rates of correct classification with sinusoidal disturbances decreased (see Table 2). When we disturbed the input data with Gaussian perturbations, the network did not got similar results (see Table 3). The correct classification rates of the network did not fall too much. The reason may be that most of the perturbed data by Gaussian perturbations were clustered around the desired value. Therefore, it indicates spiking neural networks have strong anti-interference abilities.
\subsection{Other three benchmarks}
To further validate robustness of classification ability of SNNs in practice, we consider the following three benchmarks with realistic significance. As XOR problem does, we first adopt the method in \cite{14} to encode continuous input variables in spike times. \emph{Specially, for a} variable $n$ with a range[$I_{min}^{n}$,...,$I_{max}^{n}$], we use $N$ neurons with Gaussian receptive fields to encode the input variable. For a neuron $i$, its center was set to $I_{min}^{n}+(2i-3)/2\cdot\{I_{max}^{n}-I_{min}^{n}\}/(N-2)$ and width $\sigma=1/\beta\cdot\{I_{max}^{n}-I_{min}^{n}\}/(N-2)$. We set $\beta=1.5$ for the remainder experiments. As for output classification, we encoded the patterns according to a winner-take-all paradigm.
\par (1) Iris dataset
\par The Iris flower data set or Fisher's Iris data set \cite{21} is a multivariate data set as a typical test case for many classification techniques in machine learning. The data set consists of 50 samples from each of three species of Iris (Iris setosa, Iris virginica and Iris versicolor). Four features are measured for each sample: the length and the width of the sepals and petals in centimetres, respectively. Based on the combination of these four features, we add above perturbations in each samples to build perturbed input sets. In this set of experiments, we implement a SNN with three layers just like for XOR problem. But each feature was encoded by 12 neurons with Gaussian receptive fields(yielding 48 encoding neurons and one threshold neuron). And the SNN consists of 10 hidden layer neurons (only one neuron of which generates negative signal PSPs) and 3 output neurons. The number of corresponding synaptic terminals is same as the last experiment(i.e.$m=16$). We train this SNN with perturbed input data or clear input data to distinguish the species from each other. The results are presented in Table 4 and Table 5.
\begin{table}
\begin{center}
\caption{The result of the correct classification with sinusoidal perturbations on Iris dataset. ROS: Rates of the correct classification without sinusoidal perturbations, RWS: Rates of the correct classification with sinusoidal perturbations.}
\label{tab:2}       
\begin{tabular}{llll}
\hline\noalign{\smallskip}
Epoches & A & ROS & RWS  \\
\noalign{\smallskip}\hline\noalign{\smallskip}
500& 0.001 & 96.50 & 95.71 \\
500& 0.01 & 94.60 & 93.84 \\
500& 0.1 & 91.73 & 88.90\\
500& 0.2 & 91.50 & 88.40 \\
500& 0.5 & 89.35 & 87.62\\
500& 0.8 & 88.50& 87.58 \\
\noalign{\smallskip}\hline
\end{tabular}
\end{center}
\end{table}
\begin{table}
\begin{center}
\caption{The result of the correct classification with Gaussian perturbations on Iris dataset. ROG: Rates of the correct classification without Gaussian perturbations, RWG: Rates of the correct classification with Gaussian perturbations.}
\label{tab:3}       
\begin{tabular}{llll}
\hline\noalign{\smallskip}
Epoches & $r^{*}$ & ROG & RWG  \\
\noalign{\smallskip}\hline\noalign{\smallskip}
750& 0.1 & 96.10 & 96.02\\
750& 0.2 & 94.80 & 94.43\\
750& 0.3 & 94.50 & 94.05\\
750& 0.4 & 91.50 & 90.13\\
750& 0.5 & 89.56 & 88.00\\
1000& 0.1 & 96.21 & 96.66\\
1000& 0.2 & 95.75 & 94.90\\
1000& 0.3 & 94.25 & 93.76\\
1000& 0.4 & 91.00 & 89.74\\
1000& 0.5 & 89.25 & 88.85\\
1500& 0.1 & 96.25 & 96.01\\
1500& 0.2 & 95.35 & 94.60\\
1500& 0.3 & 91.27 & 90.86\\
1500& 0.4 & 91.08 & 90.67\\
1500& 0.5 & 89.21 & 89.01\\
\noalign{\smallskip}\hline
\end{tabular}
\end{center}
\end{table}

\par (2) Wisconsin breast cancer (Original) dataset
\par The breast cancer (Original) dataset \cite{22} is from the University of Wisconsin Hospitals and contains 699 case entries, divided into benign and malignant cases. Each case has 9 measurements and each measurement is assigned an integer between 1 and 10, with larger numbers indicating a greater likelihood of malignancy. In our experiments, we encoded each measurement with 7 equally spaced neurons covering the input range. We set 15 hidden layer neurons and 2 output layer neurons. The results are presented in Table 6 and Table 7.

\begin{table}
\begin{center}
\caption{The result of the correct classification with sinusoidal perturbations on Wisconsin breast cancer (Original) dataset. ROS: Rates of the correct classification without sinusoidal perturbations, RWS: Rates of the correct classification with sinusoidal perturbations.}
\label{tab:2}       
\begin{tabular}{llll}
\hline\noalign{\smallskip}
Epoches & A & ROS & RWS  \\
\noalign{\smallskip}\hline\noalign{\smallskip}
1500& 0.001 & 97.50 & 97.60\\
1500& 0.01 & 97.34 & 97.20\\
1500& 0.1 & 95.60 & 95.53\\
1500& 0.2 & 95.50 & 93.80\\
1500& 0.5 & 96.02 & 94.84\\
1500& 0.8 & 93.56 & 91.68\\
\noalign{\smallskip}\hline
\end{tabular}
\end{center}
\end{table}
\begin{table}
\begin{center}
\caption{The result of the correct classification with Gaussian perturbations on Wisconsin breast cancer (Original) dataset. ROG: Rates of the correct classification without Gaussian perturbations, RWG: Rates of the correct classification with Gaussian perturbations.}
\label{tab:3}       
\begin{tabular}{llll}
\hline\noalign{\smallskip}
Epoches & $r^{*}$ & ROG & RWG  \\
\noalign{\smallskip}\hline\noalign{\smallskip}
1000& 0.1 & 95.75 & 96.06\\
1000& 0.2 & 95.85 & 94.60\\
1000& 0.3 & 94.75 & 94.86\\
1000& 0.4 & 92.00 & 91.96\\
1000& 0.5 & 91.57 & 91.17\\
1500& 0.1 & 97.40 & 97.52\\
1500& 0.2 & 97.13 & 96.59\\
1500& 0.3 & 95.57 & 93.86\\
1500& 0.4 & 96.03 & 95.45\\
1500& 0.5 & 93.54 & 91.60\\
\noalign{\smallskip}\hline
\end{tabular}
\end{center}
\end{table}
\par (3) StatLog landsat dataset
\par To test the robustness of SNNs on a larger dataset, we investigated the Landsat dataset as described in the StatLog survey of machine learning algorithms [23]. This dataset consists of a training set of 4435 cases and a test set of 2000 cases and contains 6 ground cover types (classes). Each sample contains the values of a 3¡Á3 pixel patch
and each pixel is described by 4 spectral bands. For classification of a single pixel, each case contains the values of a 3¡Á3 pixel patch, with each pixel described by 4 spectral bands, totaling 36 inputs per case. For each band, we used the average value of corresponding bands of 9 pixels as a new band of a pixel. Then the case was represented with one average pixel and
each separate band of it was encoded with 25 neurons. In this set of experments The results obtained by the Statlog survey are summarized in Table 8 and Table 9.
\begin{table}
\begin{center}
\caption{The result of the correct classification with sinusoidal perturbations on StatLog landsat dataset. ROS: Rates of the correct classification without sinusoidal perturbations, RWS: Rates of the correct classification with sinusoidal perturbations.}
\label{tab:2}       
\begin{tabular}{llll}
\hline\noalign{\smallskip}
Epoches & A & ROS & RWS  \\
\noalign{\smallskip}\hline\noalign{\smallskip}
6000& 0.001 & 85.50 & 85.61\\
6000& 0.01 & 85.17 & 84.80\\
6000& 0.1 & 85.00 & 84.90\\
6000& 0.2 & 85.21 & 85.20\\
6000& 0.5 & 84.50 & 82.32\\
6000& 0.8 & 83.10 & 82.04\\
\noalign{\smallskip}\hline
\end{tabular}
\end{center}
\end{table}
\begin{table}
\begin{center}
\caption{The result of the correct classification with Gaussian perturbations on StatLog landsat dataset. ROG: Rates of the correct classification without Gaussian perturbations, RWG: Rates of the correct classification with Gaussian perturbations.}
\label{tab:3}       
\begin{tabular}{llll}
\hline\noalign{\smallskip}
Epoches & $r^{*}$ & ROG & RWG  \\
\noalign{\smallskip}\hline\noalign{\smallskip}
6000& 0.1 & 85.30 & 85.02\\
6000& 0.2 & 85.07 & 84.45\\
6000& 0.3 & 83.50 & 82.83\\
6000& 0.4 & 83.46 & 83.15\\
6000& 0.5 & 81.56 & 81.00\\
7500& 0.1 & 85.60 & 85.62\\
7500& 0.2 & 85.00 & 84.75\\
7500& 0.3 & 84.50 & 83.80\\
7500& 0.4 & 82.58 & 82.15\\
7500& 0.5 & 81.80 & 80.97\\
\noalign{\smallskip}\hline
\end{tabular}
\end{center}
\end{table}
\par From the above results, we could conclude that the application of SNNs with SpikeProp algorithm on temporally encoded versions of benchmark problems yields a satisfactory robustness for sinusoidal and Gaussian perturbations.
\section{Conclusions}
In this work, the robustness of the classification ability of SNNs has been investigated by disturbing input signals with different perturbation methods not only for the classical XOR problem but also for some other complicated realistic problems. From the experiments results, it can be concluded that nevertheless perturbations will affect the classification ability of SNNs, the classification ability does not decrease dramatically and the networks have a certain anti-interference capability.



\begin{thebibliography}{}
%
%
\bibitem{1}
Rochester, N., J.H. Holland, L.H. Habit, and W.L, Duda,Tests on a cell assembly theory of the action of the brain, using a large digital computer. IRE Transactions on Information Theory,2,80-93(1956)
\bibitem{2}
Thorpe S, Fize D, Marlot C, Speed of processing in the human visual system. Nature,381,520-522 (1996)
\bibitem{3}
Werbos, P.J,Beyond Regression: New Tools for Prediction and Analysis in the Behavioral Sciences. PhD thesis, Harvard University, (1974)
\bibitem{4}
Wolfgang Maass,Networks of spiking neurons: The third generation of neural network models.Neural Networks,10,1659-1671(1997)
\bibitem{5}
M Riesenhuber, T Poggio,Hierarchical models of object recognition in cortex. Nature neuroscience,2,1019-1025(1999)
\bibitem{6}
Fukushima, K.,Neocognitron: A self-organizing neural network model for a mechanism of pattern recognition unaffected by shift in position. Biological Cybernetics,36,93-202(1980)
\bibitem{7}
A. Graves, M. Liwicki, S. Fernandez, R. Bertolami, H. Bunke, J. Schmidhuber,A Novel Connectionist System for Improved Unconstrained Handwriting Recognition. IEEE Transactions on PAMI,31,855-868(2009)
\bibitem{8}
D.V. Buonomano, M. Merzenich. A neural network model of temporal code generation and position-invariant pattern recognition. Neural Comput,11,103-116(1999)
\bibitem{9}
Bienenstock,E, A model of neocortex.Network:Computation in Neural Systems,6,179-124(1995)
\bibitem{10}
Vogl, T.P., Mangis, J.K., Rigler, A.K., Zink, W.T., Alkon, D.L., Accelerating the convergence of the back-propagation method,Biol.Cybern.,59,257-263(1988)
\bibitem{11}
Ghosh-Dastidar, S., Adeli, H.,Improved spiking neural networks for EEG classification and epilepsy and seizure detection,Integr.Comput.-Aid E.,14,187-212(2007)
\bibitem{12}
Maass,W., Noisy spiking neurons with temporal coding have more computational power than sigmoidal neurons,In: Advances in Neural Information Processing Systems,MIT Press,Cambridge,USA,9,211-217 (1997)
\bibitem{13}
Wolfgang Maass.Noise as a Resource for Computation and Learning in Networks of Spiking Neurons.Proceedings of the IEEE,102,860-880(2014)
\bibitem{14}
Sander M.Bothe, Joost N.Kok, Han La Poutre. Error-backpropagation in temporally encoded networks of spiking neurons. Neurocomputing 48,17-37(2002)
\bibitem{15}
Gerstner,W., Kistler,W., Spiking Neuron Models.Cambridge University Press,England (2002)
\bibitem{16}
Jie Yang, Wenyu Yang, Wei Wu. A remark on the error-backpropagation learning algorithm for spiking. Applied Mathematics Letters,25,1118-1120(2012)
\bibitem{17}
Haibo Bao, Jinde Cao. Delay-distribution-dependent state estimation for discrete-time stochastic neural networks with random delay. Neural Networks,24,19-28(2011)
\bibitem{18}
Mo Y Z, Ding M Z,Yu J M. Stability analysis of nonlinear cellular neural networks with time-varying delay. J Chongqing Univ Nat Sci Ed,22,817-822(2010)
\bibitem{19}
Li Y B, Wang R L. Stability of Reaction-diffusion Hopfield Neural Networks with S-type Distributed Delays. J Harbin Univ Nat Sci Ed, 15,63-66(2010)
\bibitem{20}
Jie Yang, Wenyu Yang, Wei Wu. A novel spiking perceptron that can solve XOR problem. Neural Network world,21,45-50(2011)
\bibitem{21}
 Fisher, R.A. The Use of Multiple Measurements in Taxonomic Problems. Annals of Eugenics,7,179-188(1936)
\bibitem{22}
W.H. Wolberg, Cancer dataset obtained from Williams H. Wolberg from the University of Wisconsin Hospitals, Madison,
1991.
\bibitem{23}
D. Michie, D.J. Spiegelhalter, C.C. Taylor (Eds.), Machine Learning, Neural and Statistical Classification, Ellis Horwood, West Sussex, 1994.
\end{thebibliography}
\end{document}